\pdfoutput=1

\documentclass[11pt]{article}

\usepackage[]{acl}

\usepackage{times}
\usepackage{latexsym}

\usepackage[T1]{fontenc}

\usepackage[utf8]{inputenc}

\usepackage{microtype}

\usepackage{adjustbox}
\usepackage{paralist}
\usepackage{amsmath}
\usepackage{array}
\usepackage{algorithm}
\usepackage{algorithmic}
\usepackage{enumitem}
\usepackage{color}

\usepackage{fancyhdr}
\title{Commonsense and Named Entity Aware Knowledge Grounded Dialogue Generation}

 \author{Deeksha Varshney$^\dagger$, \quad Akshara Prabhakar$^\ddagger$\thanks{\hspace{4pt}Work done during an internship at IIT Patna}\hspace{4pt}, \quad Asif Ekbal$^\dagger$\\
$^\dagger$Department of Computer Science and Engineering,\\ Indian Institute of Technology Patna, India\\
$^\ddagger$Department of Information Technology,\\
National Institute of Technology Karnataka, Surathkal\\
{\tt \small $\{$1821cs13,asif$\}$@iitp.ac.in} \\
{\tt \small akshblr555@gmail.com} \\
}

\begin{document}
\maketitle

\begin{abstract}
Grounding dialogue on external knowledge and interpreting linguistic patterns in dialogue history context, such as ellipsis, anaphora, and co-references is critical for dialogue comprehension and generation. In this paper, we present a novel open-domain dialogue generation model which effectively utilizes the large-scale commonsense and named entity based knowledge in addition to the unstructured topic-specific knowledge associated with each utterance. We enhance the commonsense knowledge with named entity-aware structures using co-references. Our proposed model utilizes a multi-hop attention layer to preserve the most accurate and critical parts of the dialogue history and the associated knowledge. In addition, we employ a Commonsense and Named Entity Enhanced Attention Module, which starts with the extracted triples from various sources and gradually finds the relevant supporting set of triples using multi-hop attention with the query vector obtained from the interactive dialogue-knowledge module. Empirical results on two benchmark dataset demonstrate that our model significantly outperforms the state-of-the-art methods in terms of both automatic evaluation metrics and human judgment. Our code is publicly available at \href{https://github.com/deekshaVarshney/CNTF}{https://github.com/deekshaVarshney/CNTF}; \href{https://www.iitp.ac.in/~ai-nlp-ml/resources/codes/CNTF.zip}{https://www.iitp.ac.in/-ai-nlp-ml/resources/ codes/CNTF.zip}. 

\end{abstract}

\section{Introduction}
Neural language models usually focus on fewer language components such as sentences, phrases, or words for text analysis. However, language acts on a much broader scale - there is frequently a central theme to a conversation, and the speakers share common information in order to comprehend one another. Information is frequently reused, however to avoid overuse, same things and persons are referred in the dialogue multiple times by using relevant expressions. A dialogue becomes coherent and speakers can understand each other when all of this information is delivered in a structured, logical, and consistent manner.

\begin{figure*}[t!]
    \centering
    \includegraphics[width=0.95\textwidth]{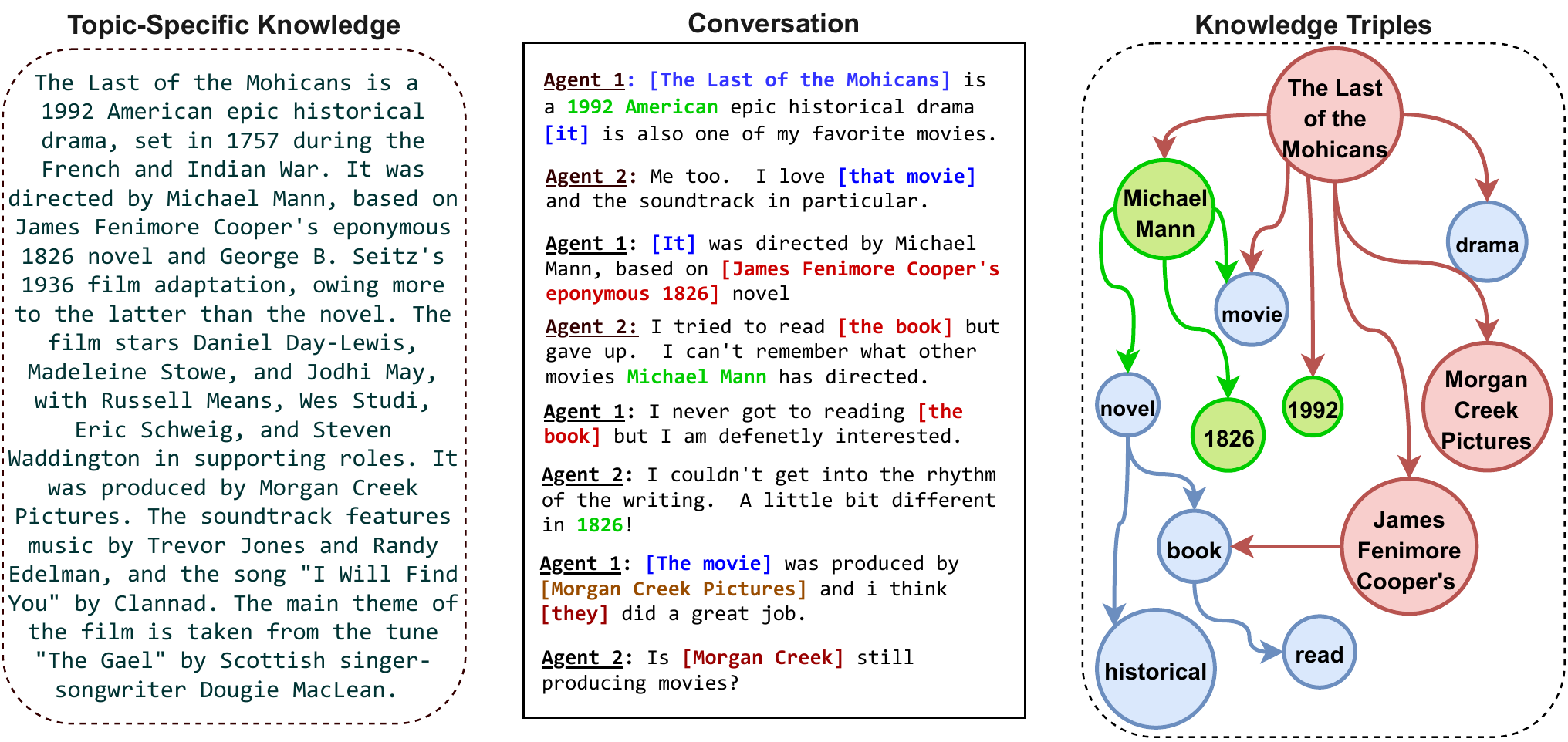}
    \caption{An example of named entity and concept based knowledge triples being used for grounding dialogues in addition to topic-specific knowledge sentences. In the conversation, various shades indicate the different co-reference clusters obtained. \textcolor{blue}{Blue} nodes correspond to the concepts obtained from ConceptNet, \textcolor{red}{red} nodes correspond to the named entities obtained from the utterances in the dialogue. Named Entities other than the ones present in co-reference chains are highlighted in \textcolor{green}{green} in the conversation.}
    \label{fig:example}
\end{figure*}

Semantic understanding of dialogues can be aided by commonsense knowledge or world facts. Additionally, as a key human language phenomena, co-reference simplifies human languages while being a significant barrier for machines to understand, particularly for pronouns, which are difficult to parse due to their weak semantic meanings \citep{ehrlich1981search}. Grounded response generation approaches \citep{ghazvininejad2018knowledge,dinan2018wizard} can provide replication of facts in open-domain settings, whereas commonsense knowledge is critical for creating successful interactions since socially constructed commonsense knowledge is the collection of contextual details that humans are expected to understand and use during a conversation. 

Despite demonstrating efficacy in empirical evaluation, past work has a few significant drawbacks. There is no explicit representation of entities, semantic relations, or conversation structures, in particular. To solve such restrictions, asking a conversation model to identify relevant structures in dialogue histories can be used to directly test the level of dialogue understanding. We focus on named entity level knowledge in this paper, and analyze references to entities in a dialogue history context. 

To ensure the generalizability of our model, we directly incorporate entities in the form of triplets, which is the most common format of modern knowledge graphs, instead of encoding it with features or rules as in conventional approaches. Take, for example, Figure \ref{fig:example}, where the dialogue consists of eight utterances. In the third utterance, to know if there exists any relation between the director ``Micheal Mann" and the movie ``The Last of the Mohicans", we need to resolve the co-reference relationship between the pronoun [It] and the entity [The Last of the Mohicans]. Using co-reference resolution, we get an important triple for the movie ``The Last of the Mohicans" \textit{viz.} \textit{(The Last of the Mohicans, RelatedTo, Micheal Mann)}. Similarly, from the second last utterance, we obtain another triple as \textit{(The Last of the Mohicans, RelatedTo, Morgan Creek Pictures)}. Thus, for instance, to generate the fourth utterance "I tried to read the book but gave up.  I can't remember what other movies \textit{Michael Mann} has directed.", it is important for the model to know that there is a relation between the concept word \textit{``movie"} and the named entities \textit{``Micheal Mann"}, \textit{``The Last of the Mohicans"}, to get a correct understanding of the dialogue context.

We create a conversational model called CNTF, \textbf{C}ommonsense, \textbf{N}amed Entity and \textbf{T}opical Knowledge \textbf{F}used neural network to generate successful responses by leveraging both topic-specific document information and using structured entity and commonsense knowledge. We first construct triples based on named entity after resolving co-references in the dialogues to enhance the already existing commonsense triples obtained from the ConceptNet \citep{speer2012representing}. We use multi-hop attention to iterate over the multi-source information. We obtain a weighted query vector from the interactive dialogue-knowledge module, which is used to query over the dialogue, topical knowledge and the corresponding triples. In each round, CNTF 
reasons on the dialogue history and knowledge sentences, using which we filter out relevant information from the dialogue context and topical knowledge. Similarly, to reason over the triples, we again iterate in multiple rounds, masking out irrelevant triples.
 
Our work makes the following contributions:

\begin{enumerate}[nolistsep]
\item We propose CNTF, a novel knowledge grounded dialogue generation model that utilizes dialogue context, unstructured textual information, and structural knowledge to facilitate explicit reasoning.

\item We enhance the commonsense triples extracted from the ConceptNet database with named entity-aware structures using co-reference resolution. 

\item We define an effective sliding window mechanism in order to remove irrelevant information from longer dialogue context and ensure efficient memory utilization. We use an interactive dialogue-knowledge module to generate a weighted query vector which captures the interactions between the conversation and the topical knowledge.

\item Through extensive qualitative and quantitative validation on publicly available datasets, we show that our model outperforms the strong baselines.

\end{enumerate}

\section{Related Work}
Sequence-to-sequence models \citep{vinyals2015neural, sutskever2014sequence} have long been used for natural language generation (NLG) tasks. Stemming off the vanilla encoder-decoder architecture - introduced initially for neural machine translation \citep{shang2015neural}, a variety of models have been developed to enhance the quality of the responses generated \citep{li2016diversity, zhao2017learning, tao2018get}; to effectively select the conversational context in multi-turn dialogues \citep{serban2016building, serban2017hierarchical, xing2017topic, zhang2019recosa}; and to model persona while conversing \citep{li2016persona, zhang2018personalizing}. Recent advances on dialogue systems aim at enhancing dialogue generation by making it more humanized by means of incorporating knowledge based on the dialogue context or from external sources, such as unstructured documents \citep{li2019incremental, qin-etal-2019-conversing} or knowledge graphs \citep{moon-etal-2019-opendialkg, tuan-etal-2019-dykgchat}.

Numerous pre-trained language models \citep{devlin-etal-2019-bert, radford2019language} have been utilized for dialogue generation \citep{edunov-etal-2019-pre, zhang2020dialogpt}. They have been extended to leverage the knowledge from the unstructured documents and other auxiliary sources via knowledge selection and various attention fusion techniques \citep{zhao-etal-2020-knowledge, cao-etal-2020-pretrained}. 
The task was explored in low-resource setting \citep{zhao2020low} using a disentangled response decoder, and the usability of language models itself as a knowledge base has also been investigated in \citet{zhao2020pre}.
An issue with language models is the noise which these introduce during knowledge selection. In order to limit the noise by generative models, term-level weighting \citep{zheng-etal-2021-knowledge} for response generation after knowledge selection were studied. \citet{zhao2020multiple} proposed a pre-training based multiple knowledge syncretic transformer that uses a single framework to integrate knowledge from multiple sources. Knowledge based end-to-end memory networks have been developed for task-oriented dialogue generation
\citep{raghu-etal-2019-disentangling, reddy2019multi, chen-etal-2019-working, wang-etal-2020-dual,varshney2021knowledge} using multi-level, working, and dynamic types of memory. In Dual Dynamic Memory Network (DDMN) \citep{wang-etal-2020-dual}, the flow of history information during conversations is dynamically tracked to retain the important parts from both dialogue and KB, using a memory manager for each. 

Prior studies \citep{young2018augmenting, zhou2018commonsense,wu2020topicka} have demonstrated the feasibility of including commonsense knowledge into the dialogue systems. Further, in ConKADI \citep{wu-etal-2020-diverse}, felicitous facts highly relevant to the context were selected and effectively integrated in the generated response by means of fusion mechanisms. Recently, co-reference resolution has been utilized for obtaining coref-informed pre-trained models \citep{ye-etal-2020-coreferential}, task-oriented dialogue generation \citep{quan2019GECOR}, and dialogue understanding \citep{zhang-etal-2021-refer}. Further, \citep{huang2021tracing} demonstrated the improvement upon explicitly incorporating co-reference information to enhance the attention mechanism for the reading comprehension task.

In this paper, we show how both structured and unstructured knowledge can be used to improve the task of document-grounded dialogue generation. We propose an effective knowledge-grounded dialogue model named CNTF, which is built with multi-source heterogeneous knowledge. Experiments on knowledge-based dialogue generation benchmark datasets, \textit{viz}. Wizard of Wikipedia and CMU\_DoG, have shown the efficacy of our proposed approach. Our method employs a large-scale named entity enhanced commonsense knowledge network as well as a domain-specific factual knowledge base to aid in the comprehension of an utterance as well as the generation of a response using a novel mutli-hop attention based model.

\begin{figure*}[ht!]
    \centering
    \includegraphics[width=0.95\textwidth]{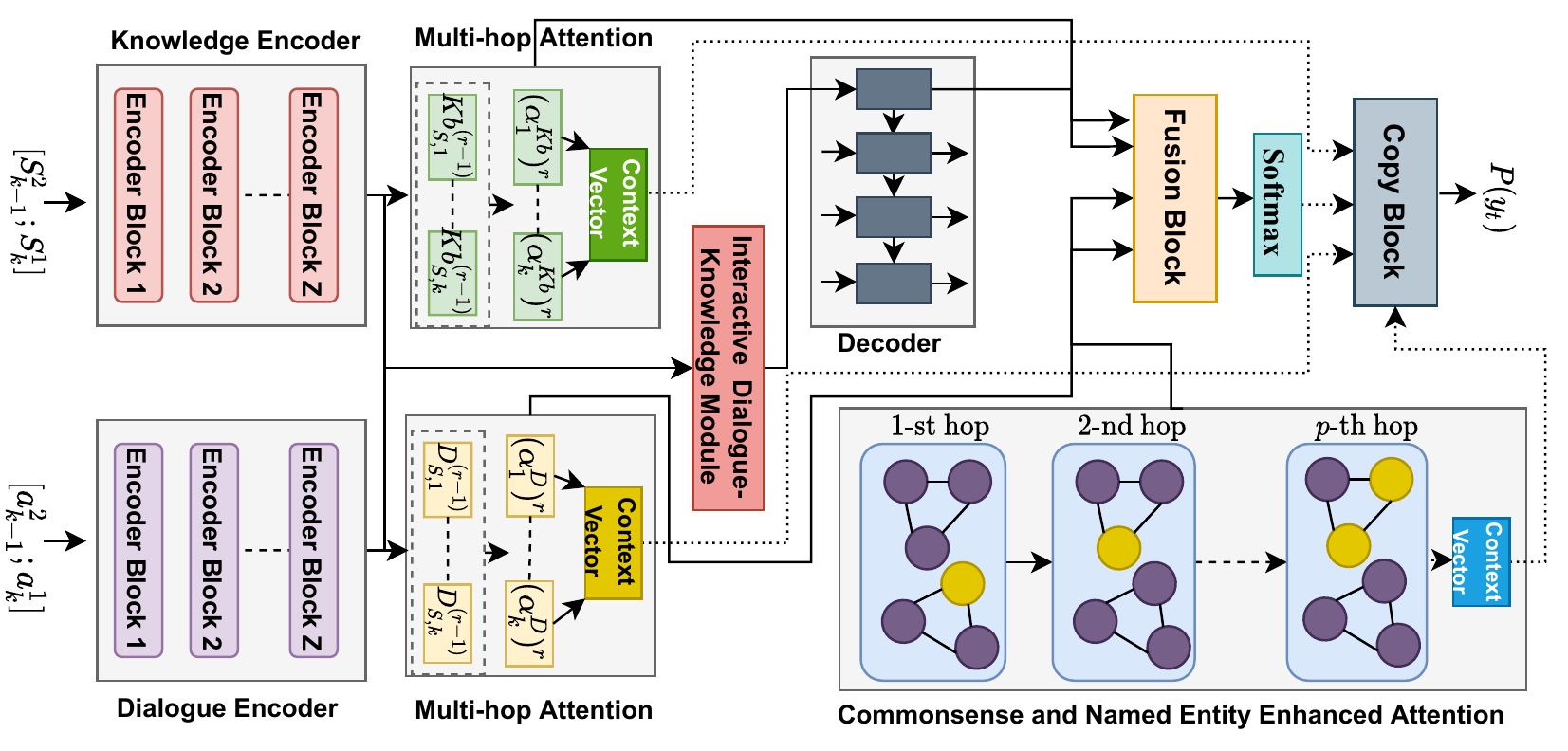}
    \caption{Proposed CNTF architecture. The dialogue encoder encodes the dialogue context in multi-turn conversation. Similarly, the knowledge encoder takes as input the document(s) associated with the utterances in the conversation. The multi-hop attention modules are used to extract relevant information from dialogue and knowledge whereas the Commonsense and Named Entity Enhanced Attention module is used to effectively incorporate the knowledge triples.}
    \label{fig:model}
\end{figure*}

\section{Methodology}
\subsection{Problem Formulation}
Formally, let $D = \{d_i\}_{i=1}^K$ denote a conversation composed of $K$ dialogue turns, where $d_i = (a^1_i, a^2_i)$ is an exchange of dialogues between the two agents. Associated with each utterance $a^1_i$ and $a^2_i$ are the relevant documents $S^1_i$ and $S^2_i$ with topic-specific knowledge. {We utilize common sense and named entity oriented knowledge by creating the set of triples $\tau = \{\tau_1, \tau_2, ..., \tau_{|\tau|}\}$, where $\tau_i$ is of the form $(head, relation, tail)$, from the following sources:
\begin{enumerate}[label=(\alph*),noitemsep,nolistsep]
    \item extracting relations from ConceptNet for every word in the utterances (if the word is a \textit{concept-word} from ConceptNet), and
    \item forming named entity based triples by using co-reference resolution method
\end{enumerate} }
For any arbitrary turn $k$, given the dialogue history $\{d_j\}_{j=1}^{k-1}$, the associated documents as well as the target document $\{S^1_j, S^2_j\}_{j=1}^{k}$, and the associated knowledge triples $\tau$, the objective is to generate an appropriate response $Y = \{y_1, y_2, ..., y_{|Y|}\}$. The architecture of CNTF is shown in Figure \ref{fig:model}. 

\subsection{Encoder}
\subsubsection{Dialogue Encoder}
The Dialogue Encoder, that keeps track of the dialogue context in multi-turn conversations, encodes the utterances turn by turn. The input at each turn is a sequence of tokens $x = (x_1, x_2, ..., x_{n})$, where n is the number of tokens. For the first turn, $a^1_1$ is fed as input, while for the subsequent turns ($j > 1$), the input is the concatenation of the previous turn's second agent's response and current turn's first agent's utterance, $[a^2_{j-1}; a^1_j]$. 
The encoder then exploits Bidirectional Encoder Representations from Transformers (BERT) \citep{devlin-etal-2019-bert} to obtain the representations $H_D = \{h_i\}_{i=1}^{n}$.

Using the dialogue representations, we maintain two different states for the dialogue, $D_S$ and $D_H$, which are both initialized with the encoder hidden states $H_D$, of the first turn. We then follow a sliding window mechanism to update both $D_S$ and $D_H$ for the succeeding turns. A window of size \textit{``l"} means we concatenate hidden states of only the previous \textit{``l-1"} turns. This helped in removing noise for longer dialogue contexts and saving memory. $D_H$ remains fixed and stores the hidden states for the dialogue context, while $D_S$ gets updated at each turn, with the goal of capturing proper history information for accurate response generation. 

\subsubsection{Knowledge Encoder}
Similarly, the Knowledge Encoder takes as input the document(s) associated with the utterances \textit{viz.} $[S^2_{j-1}; S^1_j]$ for turn $j > 1$, else $S_1^1$ for the first turn, truncated to a max token count of 400. We then again employ a BERT model and obtain the encoded features $H_{Kb} = \{h_i\}_{i=1}^{m}$, where $m$ is the number of tokens in the document(s).
To incorporate the external topic-specific knowledge effectively, we have knowledge states ${Kb}_S$ and ${Kb}_H$. Similar to the dialogue states $D_S$ and $D_H$, these are initialized with the hidden states $H_{Kb}$ of the relevant documents associated with each utterance. Unlike the sliding window mechanism used for the dialogue states for the upcoming turns, ${Kb}_S$ and ${Kb}_H$ store only the current turn's hidden states obtained from the BERT based knowledge encoder.

\subsection{Multi-hop Attention}
\label{sec:ddma}
We adopt the dual and dynamic graph attention mechanism \citep{wang-etal-2020-dual} to mimic human’s step-by-step exploring and reasoning behavior. In each step, we assume that the dialogue and knowledge states have some information to disseminate. At each hop $r$, we compute an attention vector $\alpha^{(r)}_t$ using the query embedding $q_t$ at the $k$-th turn using $D_S^{(r-1)}$ at time step $t$. $D_{H,k,t}$ and the attention scores are used to obtain the context representation $c^{(r)}_t$.
\begin{gather}
\label{eq:1}
    \alpha^{(r)}_{k,t} = softmax(e_{k,t}) \\
    e_{k,t} = ({{v}_{1}^{(r)}})^{'} \text{tanh}({W}^{(r)}_{1} q_{k,t} + {W}^{(r)}_{2} D^{(r-1)}_{S,k,t}) \\
    {c}^{(r)}_{k,t} =\sum^{K}_{j=1} a^{(r)}_{k,t} D_{H,k,t}
\end{gather}
 where ${{v}_{1}^{(r)}}$ , ${{W}_{1}^{(r)}}$ and ${{W}_{2}^{(r)}}$ are the learnable parameters.

$D_S$ is updated using the \textit{forget} and \textit{add} operations. 
To find more details on updating $D_S$ refer to Appendix \ref{sec:method}.

\subsection{Constructing Named Entity based Triples using Co-reference Resolution}
\label{sec:coref}
To add more useful links to the already existing commonsense triples, we use the co-reference chains and named entities extracted from the dialogues.
Firstly, we use AllenNLP co-reference resolution module to identify co-reference chains in the dialogue. For example, in the dialogue shown in Fig. \ref{fig:example}, using the first co-reference chain: \textbf{[The Last of the Mohicans: it, that movie, It]} we rewrite the dialogue with resolved mentions in the utterances as: \textit{``[The Last of the Mohicans] is a 1992 American epic historical drama [The Last of the Mohicans] is also one of my favorite movies. Me too.  I love [The Last of the Mohicans] and the soundtrack in particular. [The Last of the Mohicans] was directed by Michael Mann, based on [James Fenimore Cooper's eponymous 1826] novel and so on"}. We then use Spacy Named Entity tagging module to recognize named entities from the augmented dialogue. Simultaneously, we also identify all the concept words using ConceptNet in the newly formed dialogue.

The new set of triples is obtained using the named entities and concepts as nodes, and the corresponding edges are built as follows:
\begin{enumerate}[label=(\alph*),noitemsep,nolistsep]
    \item between every pair of named entities that appear in the same dialogue, and
    \item between a named entity node and other concepts within the same dialogue.
\end{enumerate} 
We may note that resolving the co-references first and then extracting named entities ensures that entities across multiple utterances are connected in a certain way. Also, we explicitly form a triplet having the \textit{RelatedTo} relation as it suits well for most of the cases because it indicates a relation between the two named entities and their different references or aliases across the conversation.

\subsection{Commonsense and Named Entity Enhanced Attention Module}
\label{sec:triple_mod}
For each dialogue, the final set of triples is composed of both commonsense and named entity based triples. We obtain triples' head and tail entity embedding from the trainable embedding layers i.e. $E = \text{emb\_layer}(\tau)$. Formally, a query is used to loop over the triple embedding and compute the attention weights at each hop $p$.
\begin{gather}
\label{eq:7}
    {\alpha}_{k,t}^{(p)} = softmax(q^{(p-1)}_t{E^{(p-1)}})
\end{gather}

Finally, the weighted context for knowledge triples, ${({c}^{T})}^{(p)}$, is obtained by weighting the current set of triple embedding, ${E}^{(p)}$ using the attention scores, $\textbf{a}^{(p)}$. A query update mechanism is used, where the query embeddings are updated using the weighted triple embeddings of the current step. 
\begin{gather}
\label{eq:8}
    {({c}_{k,t}^{T})}^p = \sum_{j=1}^{n}\textbf{a}_{k,t}^p E^{p} \\
    q^{{p}}_t = q^{p-1}_t + {({c}_{k,t}^{T})}^p
\end{gather}

\subsection{Decoder}

\subsubsection{Interactive Dialogue-Knowledge Module}
\label{sec:DKI}
As each utterance is linked to topic-specific unstructured knowledge, we employ an interactive mechanism to attend to both the dialogue and the knowledge sentences. We can improve information extraction from dialogue as well as knowledge hidden states for response generation by using the encoded weighted dialogue context as the initial query vector $q_t$. To obtain the \textit{weighted dialogue context} $WH_D$, we apply the multi-hop attention as described in Section \ref{sec:ddma} between $H_D$ and $H_K$ which are the hidden states received from the dialogue and knowledge encoder, respectively.

We use a GRU based decoder to generate responses word by word, and initialize the initial hidden states of the decoder with $WH_D$. Then at time step $t$, the decoder state $s_t$ can be updated as
\begin{gather}
\label{eq:4}
    s_t = \text{GRU}(e(y_{t-1}), s_{t-1}) 
\end{gather}

where $e(y_{t-1})$ is the embedding of the previous word $y_{t-1}$. Here, $s_t$ is regarded as the updated query vector, which is used to attend to the dialogue, topic-specific knowledge and the structured knowledge triples, and obtain the weighted context, knowledge and triple representation as ${c}^{D}_t$, ${c}^{K}_t$ and ${c}^{T}_t$, respectively. 

\subsubsection{Fusion Block}
The probability distribution over the vocabulary $P_g(y_t)$ words is obtained by fusing ${c}^{D}_t$, ${c}^{K}_t$, ${c}^{T}_t$ and the decoder state, $s_t$, and then passing them through a softmax layer.
\begin{gather}
\label{eq:9}
    P_g(y_t) = \text{softmax}({W}_{5}[s_t; {c}^{D}_t; {c}^{K}_t; {c}^{T}_t])
\end{gather}
where ${W}_{5}$ is a trainable parameter.

\subsubsection{Copy Block}
In particular, a word at time step $t$ is either generated from the vocabulary or copied from either the dialogue history, knowledge history, or using entities from the triples. Following the copy mechanism \citep{gulcehre2016pointing}, the attention scores are viewed as the probability to form the copy distribution. We use the attention score $\alpha^D_{k,t}$ of the dialogue and $\alpha^{Kb}_{k,t}$ of the unstructured knowledge at the last round \textit{viz.} $P_D(y_t=w)$ = $\sum_{tj:w_{tj}=w} \alpha^D_{k,t}$; $P_{Kb}(y_t=w)$ = $\sum_{tj:w_{tj}=w} \alpha^{Kb}_{k,t}$. The copy distribution over the triples is given by $P^T(y_t=w)$ = $\sum_{tj:w^t_j=w}\alpha^T_{k,t}$. We use the soft gates $g_1$, $g_2$ and $g_3$ to control whether a word is generated from the vocabulary or it is being copied by combining $P_g(y_t)$, $P_D(y_t)$, $P_{Kb}(y_t)$, and $P_T(y_t)$:
\begin{gather}
g_1 = \text{Sigmoid}({W}_{8}[s_t; c^D_t] + b_2) \\
P_{kn}(y_t) = g_1P_g(y_t) + (1- g_1)P_D(y_t) \\
g_2 = \text{Sigmoid}({W}_{9}[s_t; c^K_t] + b_3) \\
P_{tp}(y_t) = g_2P_{Kb}(y_t) + (1-g_2)P_{kn}(y_t) \\
g_3 = \text{Sigmoid}({W}_{10}[s_t; c^T_t] + b_4) \\
P(y_t) = g_3P_{T}(y_t) + (1-g_3)P_{tp}(y_t) 
\end{gather}
where, ${W}_{8}$, ${W}_{9}$, ${W}_{10}$ are the parameters to be learned.

Therefore, the decoder loss is the cross-entropy between the predicted distribution $P(y_t)$ and the reference distribution, $p_t$, denoted as $Loss = - {\sum}{p_t}log(P(y_t))$.

\section{Datasets and Experimental Setup}
In this section, we present the details of the datasets and the other experimental setups. Implementation details can be found in Appendix \ref{sec:imp}.

\subsection{Dataset Description}

\subsubsection{Knowledge Grounded Dialogue Dataset}
We test our proposed technique on two knowledge-grounded dialogue generation benchmark datasets, \textit{viz.} Wizard of Wikipedia \citep{dinan2018wizard} and CMU Document Grounded Conversations \citep{zhou2018dataset}. The WoZ and CMU\_DoG datasets consist of approximately $\approx$ 22K and $\approx$ 4K dialogs, respectively, covering more than 1,365 and 90 topics. 
The datasets are summarized in Appendix \ref{sec:datasets}. The statistics of the datasets are shown in Table \ref{tab:dataset_details} of the Appendix.

\subsubsection{Commonsense Knowledge Base}
We use ConceptNet, an open-domain repository of commonsense knowledge. It includes the relationships between concepts that are commonly used in everyday situations, such as "Mango is a fruit." This function is desirable in our experiments because it is critical to be able to identify the informal relationships between common concepts in an open-domain conversation setting. We remove triples containing multi-word entities when filtering words based on dataset vocabulary, and $147,676$ triples were retained with $27,468$ entities and $44$ relations for Wizard of Wikipedia dataset. For CMU\_DoG dataset, we have a total of $14,689$ entities, $74,485$ triples and $42$ relations.

\begin{table*}[t!]
\begin{center}
\begin{adjustbox}{max width=1.0\textwidth}
\renewcommand{\arraystretch}{1.3}
\setlength\tabcolsep{1.2pt}
\begin{tabular}{|>{\centering\arraybackslash}m{2.8cm}|>{\centering\arraybackslash}m{2.4cm}|>{\centering\arraybackslash}m{2.4cm}|>{\centering\arraybackslash}m{2.5cm}|>{\centering\arraybackslash}m{2.5cm}|>{\centering\arraybackslash}m{2.5cm}|>{\centering\arraybackslash}m{2.5cm}|>{\centering\arraybackslash}m{2.5cm}|>{\centering\arraybackslash}m{2.0cm}|>{\centering\arraybackslash}m{2.0cm}|>{\centering\arraybackslash}m{2.1cm}|>{\centering\arraybackslash}m{1.9cm}|>{\centering\arraybackslash}m{1.8cm}|>{\centering\arraybackslash}m{1.8cm}|>{\centering\arraybackslash}m{1.9cm}|>{\centering\arraybackslash}m{1.9cm}|>{\centering\arraybackslash}m{1.9cm}|>{\centering\arraybackslash}m{1.9cm}|>{\centering\arraybackslash}m{1.9cm}|}
\hline
\multicolumn{1}{|c|}{} & \multicolumn{6}{|c|}{\textbf{Wizard of Wikipedia}} & \multicolumn{6}{|c|}{\textbf{CMU\_DoG}} \\
\hline
\textbf{Models} & \textbf{PPL \textit{(Seen/Unseen)}}  & \textbf{F1\% \textit{(Seen/Unseen)}} & 
\textbf{BLEU-4 \textit{(Seen/Unseen)}} & \textbf{Embedding Average \textit{(Seen/Unseen)}} & \textbf{Vector Extrema
\textit{(Seen/Unseen)}} & \textbf{Greedy Matching
\textit{(Seen/Unseen)}}  & \textbf{PPL}  & \textbf{F1\%} & 
\textbf{BLEU-4} & \textbf{Embedding Average} & \textbf{Vector Extrema} & \textbf{Greedy Matching} \\
\hline
TMN &  { 66.5 / 103.6 } & { 15.9 / 14.3 } &
{ 0.017 / 0.009 } & { 0.844 / 0.839  } &  { 0.427 / 0.408  } & { 0.658 / 0.645 } &  { 75.2 } & { 9.9 }
& { 0.007 } & { 0.789  } &  { 0.399  } & { 0.615 }  \\  

ITDD &  { 17.8 / 44.8 } & { 16.2 / 11.4 }&
{ 0.025 / 0.011 } & { 0.841 / 0.826  } &  { 0.425 / 0.364  } & { 0.654 / 0.624 } &  { 26.0 } & { 10.4 }& 
{0.009 } & { 0.748  } &  { 0.390  } & { 0.587 }  \\

$\text{DialogGPT}_{\text{finetune}}$ &  \textbf{{ 16.2 / 20.4 }} & { 19.0 / 17.6 }& 
{ 0.023 / 0.017 } & { 0.871 / 0.869  } &  { 0.461 / 0.451  } & { 0.683 / 0.674 } &  \textbf{{ 15.9 }} & { 13.7 }& 
{ 0.015 } & { 0.812  } &  { 0.430  } & { 0.641 }  \\

{DRD} & { 19.4 / 23.0 } & { 19.3 / 17.9 }
& { 0.044 / 0.037 } & { 0.864 / 0.862  } &  { 0.455 / 0.444  } & { 0.679 / 0.671 } & { 54.4 } & { 10.7 }&
{ 0.012  } &  { 0.809  } & { 0.413 } & { 0.633 } \\

{ConKADI} & { 89.4 / 93.0 } & { 13.3 / 15.9 }
& { 0.016 / 0.014 } & { 0.726 / 0.662  } &  { 0.355 / 0.324  } & { 0.599 / 0.601 } & { 84.4 } & { 8.7 }&
{ 0.006  } &  { 0.768  } & { 0.326 } & { 0.600 } \\

{KnowledGPT} & { 19.2 / 22.3}& { 22.0 / 20.5 }& 
{ 0.058 / 0.047   } &  { 0.872 / 0.870  } & { 0.463 / 0.452 } & { 0.682 / 0.674 } & { 20.6 }& { 13.5 }& 
{  -  } &  { 0.837  } & { 0.437 } & { 0.654 } \\
\hline
\textbf{CNTF} & { 24.4 / 28.6  }& {{ 32.5} / 31.4 }& 
\textbf{ 0.119} / 0.110   &  {0.911} / 0.910 & {{ 0.577
}/ 0.570 } & {{ 0.758} / 0.752} & {  46.0  } & {{14.6}} & 
\textbf{{0.018}} &  \textbf{{0.882} }& \textbf{{0.518}} & \textbf{{0.708  }} \\
\hline

CNTF-DKIC & { 24.3 / 28.5 }& \textbf{ 33.1 / 32.9} & 
{ 0.118 / \textbf{0.117}  } &  \textbf{{0.913 / 0.913  }} &\textbf{{ 0.582 / 0.581}} & \textbf{{ 0.761 / 0.758 }} & {  44.5  } & \textbf{{15.1}} & 
{0.018} &  {0.882} & {0.518} &{0.708  } \\

{CNTF-DKI} & { 26.8 / 31.8 }& { 32.4 / 31.5 }& 
{ 0.114 / 0.110  } &  { 0.911 / 0.912  } & { 0.576 / 0.575 } & { 0.758 / 0.754 } &  { 45.3   }& {14.2} & 
{ 0.015  }&  { 0.881   } & {0.514} & {0.707} \\
{CNTF-DK} & { 25.9 / 31.1  }& { 30.9   / 29.8 }& 
{ 0.105 / 0.101  } &  { 0.909 / 0.909  } & { 0.567 / 0.564 } & { 0.752 / 0.746 }  & {  45.9  }& {  14.1  }&
{  0.014   } &  {  0.880  } & {  0.505  } & {  0.700  }\\
{CNTF-D} & { 47.5 / 96.3 }& { 15.3 / 13.5 }& 
{ 0.022 / 0.015  } &  { 0.884 / 0.883  } & { 0.456 / 0.440 } & { 0.689 / 0.679 } & {  47.9  }& { 11.8 }& 
{    0.013 } &  {   0.880 } & {  0.492  } & {  0.693  } \\

\hline
\end{tabular}
\end{adjustbox}
\end{center}

\caption{\label{tab:auto_results} Automatic evaluation results marked in bold fonts indicate the best outcome for the measure and improvement over the best baseline, and is statistically significant (t-test with p-value at 0.05 significance level). The scores on the ablation models are shown in the last four rows. The values for baseline models are derived from \citep{zhao-etal-2020-knowledge} and \citep{zhao2020pre}. (-) indicates that the value was not reported.}

\end{table*}

\begin{table*}[ht!]
\begin{center}
\small
\begin{adjustbox}{max width=0.70\textwidth}
\renewcommand{\arraystretch}{1.3}
\setlength\tabcolsep{1.2pt}
\begin{tabular}{|>{\centering\arraybackslash}m{1.9cm}|>{\centering\arraybackslash}m{1.9cm}|>{\centering\arraybackslash}m{1.9cm}|>{\centering\arraybackslash}m{2.0cm}|>{\centering\arraybackslash}m{2.0cm}|>{\centering\arraybackslash}m{2.0cm}|>{\centering\arraybackslash}m{2.0cm}|}
\hline
\textbf{Models} & \textbf{Fluency \textit{(Seen/Unseen)}}  & \textbf{Adequacy \textit{(Seen/Unseen)}} &  \textbf{Knowledge Existence \textit{(Seen/Unseen)}} &  \textbf{Knowledge Correctness \textit{(Seen/Unseen)}} &  \textbf{Knowledge Relevance \textit{(Seen/Unseen)}} & \textbf{Kappa \textit{(Seen/Unseen)}} \\
\hline
TMN &  { 1.314 / 1.197}   &  { 	1.262 /   0.934}  & {	1.046 / 	0.811 }& {1.005 / 0.691 }& {0.867  / 	0.487 }  & { 0.931 / 0.888} \\ 
ITDD &  { 1.135	/ 1.290}   &  {0.545 / 	0.965 }  & {0.515 / 0.382 }  & {0.301 / 0.188 } & {0.184  / 0.101 }  & {0.940 / 0.930}\\   
KnowledGPT &  \textbf{1.813 / 1.817}   &   {1.568 / \textbf{1.556}}   &  {1.493 / 	1.139}   &  {1.430 / 1.390}   &  {1.172 / 1.040}   &  {0.810 / 0.811}   \\  

\hline
\textbf{CNTF} &  {1.561 / 1.554}   &   {\textbf{1.647} / 1.469}   & \textbf{1.653 / 1.285}   & \textbf{1.770 / 1.422}   & \textbf{1.732 / 1.376}   &  {0.830 / 0.869}   \\  
\hline
{Gold Response} &  {1.865 / 1.883}   &   {1.891 / 1.883}   & {1.825 / 1.864}   & {1.908 / 1.916}   & {1.903 / 1.904}   &  {0.890 / 0.854}  \\
\hline
\end{tabular}
\end{adjustbox}
\end{center}

\caption{\label{tab:h_results} Human assessment results for the baseline and proposed model on WoZ dataset. Bolded results of the proposed model against the baselines are statistically significant using t-test at 0.05\% significance level. }

\end{table*}

\subsection{Baselines}
We use the following models as the baselines:

\begin{inparaenum}
\item \textbf{Transformer Memory Network (TMN)} \citep{dinan2018wizard}: To encode dialogue, a shared transformer-based encoder is used. After knowledge selection, memory networks are used to re-encode the dialogue information. Finally, a transformer decoder is used to decode the responses.

\item \textbf{$\text{DialogGPT}_{\text{\textit{finetune}}}$}\citep{zhao2020pre}: It utilises a DialoGPT (345M) model fine-tuned on training examples from the Topical Chat dataset to determine whether the pre-trained models can serve as knowledge bases for open-domain dialogue generation.

\item \textbf{Incremental Transformer with Deliberation Decoder (ITDD)} \citep{li2019incremental}: It uses an incremental transformer-based model to encode utterances and documents and a deliberation decoder to decode responses.

\item \textbf{Disentangled Response Decoder (DRD)} \citep{zhao2019low}: It is made up of three modules: a language model, a context processor, and a knowledge processor for decoding responses. The response decoder is broken down into independent components in this case to investigate knowledge-based dialogue generation.

\item \textbf{ConKADI} \citep{wu-etal-2020-diverse}: It includes a Felicitous Fact mechanism to help the model focus on knowledge facts that are highly significant; additionally, two techniques, Context-Knowledge Fusion and Flexible Mode Fusion, are proposed to assist ConKADI in integrating the knowledge information.

\item \textbf{KnowledGPT} \citep{zhao-etal-2020-knowledge}:
This model implements response generation by combining a pre-trained language model with a knowledge selection module, and it intends to jointly optimize knowledge selection and response generation with unlabeled dialogues using an unsupervised approach.

\end{inparaenum}

\subsection{Evaluation Metrics}
To evaluate the predicted responses, we choose BLEU \citep{papineni2002bleu}, PPL, F1 and Embedding-based metrics \citep{liu2016not}. For human evaluation, we use fluency, adequacy, knowledge existence, knowledge correctness and knowledge relevance. Appendix \ref{sec:metrics} provides more information on these metrics.

\section{Results and Analysis}
\label{sec:result}

\subsection{Results of Automatic Evaluation}
Table \ref{tab:auto_results} shows the results on automatic evaluation metrics on Wizard of Wikipedia and CMU\_DoG datasets. On WoZ, CNTF gives a significant rise of 48\% on Test Seen and 53\% on Unseen in F1 score and around two times more on both Seen and Unseen, in terms of BLEU-4, compared to the strongest baseline, KnowledGPT. On CMU\_DoG too, where the average turn length is roughly 2.5 times that of WoZ, CNTF surpasses the previous best on F1 and BLEU-4 by 8\% and 20\% respectively. Hence, CNTF achieves new state-of-the-art on both datasets.

Existing models struggle to generate engaging responses for dialogues based on new topics that were not encountered during the training phase, which most likely explains the observed low performance on Test Unseen. On the contrary, CNTF is capable of capturing the dialogue context and effectively utilizing external commonsense knowledge and parse the implicit mentions made to various entities through the conversation to produce accurate responses, as evidenced by the magnitude of improvement achieved. On embedding-based metrics, all three measures have significantly improved, demonstrating the efficacy of our methodology. Comparison to more baseline models can be found in Appendix \ref{sec:auto_r}.

\begin{table*}[ht!]
\small
\begin{center}

\begin{adjustbox}{max width=1.0\textwidth}
\renewcommand{\arraystretch}{1.4}
\setlength\tabcolsep{1.3pt}
\begin{tabular}{|>{\centering\arraybackslash}p{2.0cm}||p{17.7cm}|}
\hline
\textbf{Utterance 1} & \textbf{Agent 1:} \textit{i love dr pepper}  \\
\textbf{Knowledge 1} & \textit{dr pepper is a carbonated soft drink marketed as having a unique flavor.}   \\
\textbf{Triples} & \textit{(pepper, RelatedTo, soda), (soda, IsA, drink), (flavor, RelatedTo, drink), (carbonated, Synonym, soda), (soft, RelatedTo, drink), (drink, RelatedTo, like)}... 
\\
\hline
\textbf{CNTF} & {yeah i love dr pepper. it is marketed by having a unique flavor.} \\

\textbf{KnowledGPT} & i love dr pepper too! it's a carbonated soft drink marketed as having a unique flavor. \\
\hline
\textbf{Utterance 2} & \textbf{Agent 2:} \textit {Oh same, it's my favorite Dr Pepper is a carbonated soft drink } \textbf{Agent 1:} \textit{oh yea? its such a good soda}  \\
\textbf{Knowledge 2} & \textit{7 up (stylized as 7 up outside the u.s.) is a brand of lemon-lime flavored, non-caffeinated soft drink.}\\
\textbf{Triples} & \textit{ (flavor, RelatedTo, drink), (pepper, RelatedTo, soda), (soda, IsA, drink), (carbonated, Synonym, soda), (soft, RelatedTo, drink), (US, RelatedTo, pepper), (great, RelatedTo, super)
} ...\\
\hline
\textbf{CNTF} & yea 7 up (stylized as 7up outside the u.s.) is a brand of lemon-line.\\
\textbf{KnowledGPT} & it was created in the 1880s by pharmacist charles alderton in waco, texas and first served around 1885. \\

\hline
\textbf{Utterance 3} & \textbf{Agent 2:} \textit{oh yeah, i also like 7-up, 5 a brand of lemon-lime \textbf{flavor}.}  \textbf{Agent 1:} \textit{oh yeah thats great too}  \\
\textbf{Knowledge 3} & \textit{the drink was created in the 1880s by pharmacist charles alderon in waco, texas and first served around 1885.}   \\
\textbf{Triples} &\textit{ \textbf{(flavor, RelatedTo, drink),} (1880s, RelatedTo, wild),(1880s, RelatedTo, created),(1880s, RelatedTo, pharmacist), \textbf{(drink, RelatedTo, pepper)} ...}
\\
\hline

\textbf{CNTF} & yeah , the \textbf{drink} was created in the 1880s by pharmacist charles alderton in waco. \\
\textbf{KnowledGPT} & yeah, 7 up was created by charles leiper grigg, who launched his st. louis based company the howdy corporation in 1920 \\
\hline
\textbf{Utterance 4} & \textbf{Agent 2:} \textit{but, dr pepper was created in 1880s by a pharmacist. isn't that wild? } \textbf{Agent 1:} \textit{yea wow it is so old}  \\
\textbf{Knowledge 4} & \textit{dr pepper was first nationally marketed in the united states in 1904, and is now also sold in europe, asia, canada, mexico, australia, and south america, as well as new zealand and south africa as an imported good.}  \\

\textbf{Triples} & \textit{\textbf{(1904, RelatedTo, pepper), (sold, RelatedTo, Europe)}, (pepper, RelatedTo, Australia), (pepper, RelatedTo, Canada)} ... \\
\hline
\textbf{CNTF} & yea dr \textbf{pepper} was first nationally sold in the us in \textbf{1904} , and is now also \textbf{sold} in \textbf{europe}.  \\
\textbf{KnowledGPT} & it was first served in waco, texas in 1885. i wonder how long it has been around? \\
\hline
\textbf{Utterance 5} & \textbf{Agent 2:} \textit{oh yeah, and it didn't make it into the us until it was first nationally marketed in the united states in 1904} \textbf{Agent 1:} \textit{thats super crazy}  \\
\textbf{Knowledge 5} & \textit{dr pepper was first nationally marketed in the united states in 1904, and is now also sold in europe, asia, canada, mexico, australia, and south america, as well as new zealand and south africa as an imported good.}  \\

\textbf{Triples} & \textit{(1904, RelatedTo, marketed), (States, RelatedTo, 1904), (United, RelatedTo, 1904), (1904, RelatedTo, US), (1904, RelatedTo, first), (first, RelatedTo, marketed) ...} \\
\hline
\textbf{CNTF} & yea dr pepper was first nationally marketed in the us in 1904 , and is now also sold in \\
\textbf{KnowledGPT} & it was formulated by a pharmacist named charles alderton in morrison's old corner drug store in waco, texas. \\
\hline
\end{tabular}
\end{adjustbox}
    
\end{center}
\caption{
\label{tab:gen_resp} {Samples from Test Seen of WoZ dataset. The gold response for the $(k)$-th example is Agent 2's utterance in the $(k+1)$-th example. The displayed knowledge is the supporting sentence for the gold response to that utterance.}}

\end{table*}


\subsection{Human Evaluation Results}
Human evaluation results are shown in Table \ref{tab:h_results}. We only compare our proposed model against KnowledGPT, ITDD and TMN on WoZ, as manual evaluation is expensive. It is clear that CNTF outperforms the baselines on both adequacy and knowledge-related criteria, demonstrating consistency with the results of automatic evaluation, and has comparable fluency performance. It is important to note that, despite providing contextually appropriate responses, KnowledGPT failed to capture the accurate knowledge associated with the input sequences, resulting in low scores. The strength of CNTF can be seen from the knowledge existence, correctness and relevance scores. This can be attributed to the fact that the multi-hop attention module and the interactive attention module incorporate the knowledge bases efficiently. The knowledge in the generated response is relevant with the contexts and is factually correct. Furthermore, the responses are more effective at exchanging information than at casual chat. The proposed model also makes good use of commonsense knowledge and named entities due to attention module as explained in Section \ref{sec:triple_mod}. All of the kappa values are greater than 0.75, indicating that the annotators agree.

In Table \ref{tab:gen_resp}, we present a few example conversations as predicted by the proposed (\textit{CNTF}) and the strongest baseline (\textit{KnowledGPT}) on Test Seen from Wizard of Wikipedia.
In utterance 3, \textit{CNTF} is able to decipher that the context of the conversation is \textit{dr. pepper} using the triple \textit{(drink, RelatedTo, pepper)} obtained using the mechanism explained in Section \ref{sec:triple_mod} unlike KnowledGPT which starts talking about \textit{7up}. Additionally, \textit{CNTF} efficiently utilizes the commonsense knowledge triples by correctly copying the entities in the triples associated with the word \textit{flavor}. As seen in the fourth utterance, the model correctly decodes the response using more detailed knowledge from the topic-specific knowledge base as opposed to KnowledGPT. Triples such as \textit{(1904, RelatedTo, pepper), (sold, RelatedTo, Europe)} which were created using Section \ref{sec:coref} have aided it in understanding the context better.

\subsection{Ablation Study}
\label{sec:abl}
To analyze the impact of the constituent modules in our model on performance (Table \ref{tab:auto_results}), we compare CNTF with the following variants:

\begin{inparaenum}[(i)] 
\item CNTF-D: This configuration only employs the dialogue encoder with multi-hop attention to demonstrate the significance of employing a knowledge encoder. This results in a 53\% decrease in F1 score on Test Seen, demonstrating the effectiveness of our knowledge module with multi-hop attention. The score reduction in CMU\_DoG is less severe because workers do not rely as heavily on external knowledge as the Wizard does, where it is highly correlated with available knowledge.
\item CNTF-DK: Interactive attention is essential for generating insightful responses while decoding the answer. We remove the Interactive Dialogue-Knowledge module, as explained in Section \ref{sec:DKI}, to demonstrate its utility. This results in a significant decrease in both BLEU and F1 scores.
\item CNTF-DKI: We conduct experiments with only the dialogue and knowledge modules, as well as the interactive module, to demonstrate the benefit of using structured knowledge in the form of triples for gauging the implicit references made throughout the conversation. We see a significant drop in scores here as well.
{\item CNTF-DKIC: This model is used to show the effectiveness of co-reference based named entity triples. We see a drop in BLEU-4 scores for the seen testset, but we see an improvement on the unseen testset by using only commonsense knowledge. This could be attributed to the fact that for unseen data, the same entities are usually not present because they have conversations on topics that are rarely seen in the training set.}
\end{inparaenum}

We may note that CNTF beats the SOTA models on every metric however due to the addition of new triples (more than 60\% increment in triples on an average for both the datasets) which may have added to noise in the model, and hence it shows 
lower scores on some metrics than CNTF-DKIC.

\subsection{Error Analysis} 
\label{sec:error_analysis}
Using the generated responses, we perform a thorough examination of our proposed model and categorize the errors it encounters as follows:

\begin{inparaenum}
\item \textbf{Repetition}: There are some instances where certain words are repeated in the generated responses. For example, \textit{Predicted response: ``i' m not sure, but it is similar to violet, violet and violet."} 

\item \textbf{Incomplete response:} As shown in the response for the last example in Table \ref{tab:gen_resp}, incomplete responses result in lower fluency scores. We discovered that the ground truth responses in the dataset are generated by copying incomplete sentences from the document knowledge. Since our model augments knowledge, it learns to produce responses in the same manner. For example: \textit{Document knowledge}: ``\textit{there is no scientifically precise definition of genius, and the question of whether the notion itself has any real meaning has long been a subject of debate, although psychologists are converging on a definition that emphasizes creativity and eminent achievement.}''; \textit{Gold Response:} ``\textit{there is no scientifically precise definition of genius}''. As can be seen, the response picked is incomplete and less fluent if it is compared to the knowledge sentence.
We have evaluated our gold responses considering this in Table \ref{tab:h_results}. We observed that the fluency score is 1.865 / 1.883 for both the test seen / unseen set. A few more error cases with examples are shown in the Appendix \ref{sec:error_ext}.

\end{inparaenum}

\section{Conclusion }
\label{sec:conclusion}
We present a Commonsense, Named Entity, and Topical Knowledge Fused neural network (CNTF) to address reasoning over multiple knowledge bases in this paper. We propose, in particular, multi-hop attention over both structured and unstructured knowledge. Unlike previous approaches in Dialog, CNTF can find relevant supporting named entities in dialogs at each step of multi-hop attention in addition to already present commonsense knowledge. We test CNTF on WoZ and CMU\_DoG and achieve excellent results. Furthermore, our analysis shows that CNTF can generate consistent results.

{In the future, we hope to expand our work to build models which include emotions for knowledge grounded dialogues. Also, to tackle repetition and incomplete response, we aim to introduce rewards functions for these factors. Currently, our model does not consider the relation attribute in our proposed framework and hence the use of “RelatedTo” relation is not really affecting the performance of the proposed approach. We aim to incorporate relation attributes for triple representations.}

\section{Ethical Declaration}
Our work relies solely on publicly available data. We followed the policies for using the data and did not violate any copyright issues.

\section*{Acknowledgement}
Asif Ekbal acknowledges the Young Faculty Research Fellowship (YFRF), supported by Visvesvaraya Ph.D. scheme for Electronics and IT, Ministry of Electronics and Information Technology (MeitY), Government of India, being implemented by Digital India Corporation (formerly Media Lab Asia).

\bibliography{anthology,custom}
\bibliographystyle{acl_natbib}

\appendix
\section{Methodology}
\label{sec:method}
To update $D_S$, we use another Gated Recurrent Unit (GRU) network to emulate the decoder at round $r$, obtaining the “intermediate” hidden states, $\Tilde{s}^{(r)}_t$. 
\begin{gather}
\label{eq:3}
    \tilde{s}^{(r)}_t = \text{GRU}(c^{(r)}_t, q_{t}) 
    \\
  \tilde{u}^{(r)}_{t} = u^{(r-1)}_{t}(1-\tilde{a}^{(r)}_{t} F^{(r)}_t) \\
  F^{(r)}_t = \text{Sigmoid}({W}^{(r)}_{3}, \tilde{s}^{(r)}_t)
  \\
  D^{(r)}_{S,t} = \tilde{u}^{(r)}_{t} + \tilde{a}^{(r)}_{t} A^{(r)}_t \\
   A^{(r)}_t = \text{Sigmoid}({W}^{(r)}_{4}, \tilde{s}^{(r)}_t)
\end{gather}
${W}_{3}^{(r)}$ and ${W}_{4}^{(r)}$ are the learnable parameters. $\tilde{a}^{(r)}_{t}$ is computed similar to the manner defined in Eq \ref{eq:1}.

\section{Datasets}
\label{sec:datasets}
Experiments are carried out on two benchmark datasets, \textit{viz.} Wizard of Wikipedia \citep{dinan2018wizard} and CMU\_DoG \citep{zhou2018dataset}. 

Wizard of Wikipedia (WoZ) is one of the most comprehensive knowledge-based conversation datasets, covering 1,365 open-domain topics. Each conversation takes place between a wizard who can retrieve knowledge about a specific topic and form a response based on it and an apprentice who is simply eager to speak with the wizard but lacks access to external knowledge. The test set is further divided into two parts: Test Seen and Test Unseen. The former contains conversations about topics that have previously been seen in the training set, whereas the latter contains conversations about topics that have never been seen in either the training or validation sets.

CMU\_DoG focuses on the movie domain, and the conversations take place between two users who both have access to the relevant documents. Every document includes information, such as the title of the film, the cast, an introduction, ratings, and a few scenes. We consider subsequent utterances by the same person as a single one. ConceptNet database can be downloaded from \href{https://conceptnet.io}{https://conceptnet.io}.

\begin{table}[ht!]
\renewcommand{\arraystretch}{1.3}
\setlength\tabcolsep{1.2pt}
\begin{adjustbox}{max width=0.50\textwidth}
\centering
\small
\begin{tabular}{|>{\centering\arraybackslash}m{2.8cm}|>{\centering\arraybackslash}m{1.0cm}|>{\centering\arraybackslash}m{1.3cm}|>{\centering\arraybackslash}m{1.2cm}|>{\centering\arraybackslash}m{1.3cm}|>{\centering\arraybackslash}m{1.0cm}|>{\centering\arraybackslash}m{1.3cm}|>{\centering\arraybackslash}m{1.0cm}|}
\hline
\multicolumn{5}{|c|}{Wizard of Wikipedia} & \multicolumn{3}{|c|}{CMU\_DoG} \\
\hline
\textbf{} & \textbf{Train} & \textbf{Valid} & \textbf{Test Seen} & \textbf{Test Unseen} & \textbf{Train} & \textbf{Valid} & \textbf{Test}  \\ 
\hline
\textbf{\#Conversation} & 18,430 &  1,948 &  965  & 968 & 3,373 & 229 & 619 \\
\hline
\textbf{\#Utterances} &   166,787 &  17,715  & 8,715  & 8,782  & 74,717 & 4,993 & 13,646 \\
\hline
\textbf{Avg. \# of Turns} & 9.0 &  9.1 &  9.0  & 9.1 & 22.2 & 21.8 & 22.0  \\
\hline
\textbf{\#Topics\/Documents}  & 1,247 &  599 &  533 & 58 & 30 & 30 & 30\\

\hline
\end{tabular}
\end{adjustbox}
\caption{\label{tab:dataset_details} Dataset Statistics }
\end{table}

\section{Implementation Details}
\label{sec:imp}
For our proposed CNTF model, we set the word embedding dimension as 300, and use GloVe word embeddings. The hidden size of GRU is sampled from $\{128, 256\}$. Both the number of rounds $R$, the number of hops $K$ are sampled from $\{2, 3\}$, and the sliding window size is sampled from $\{1,2\}$. We use the ADAM optimizer \citep{kingma2014adam} whose learning rate is fixed to 0.0005 and set the beam size to 4, while decoding the responses. We truncate utterances to a max token count of 200 and knowledge base to 400. To handle the long-text knowledge base of CMU\_DoG, for every utterance and knowledge sentence we compute a TF-IDF vector. We then compute the cosine similarity between an utterance and every sentence in the knowledge base and retain the top-2 knowledge sentences, similar to the procedure adopted in Enriched Topical Chat dataset \citep{gopalakrishnan2019topical}. The conversation and knowledge base vocabulary is shared and comprises of 30,004 words, while common sense vocabulary is maintained separately. We choose batch size as 2 and 8 for CMU\_DoG and Wizard of Wikipedia, respectively, for training the models. There are roughly 83M parameters for our model when trained on Wizard of Wikipedia, and 38M on CMU\_DoG, the difference in size is due to the vocabulary variation. These are much lesser than large pre-trained models which have much greater parameters (KnowledGPT which uses GPT-2). It is trained for 10-15 epochs. We choose the best model when the loss on the validation set does not decrease. The variances of the results are at most 1e-3 after three runs with random initialization for each method, and they have no effect on the trend. We have adapted the code framework from \href{https://github.com/siat-nlp/DDMN}{DDMN} \citep{wang-etal-2020-dual}. We have used GeForce GTX 1080 Ti as the computing infrastructure.
We used the AllenNLP co-reference resolution module (\href{https://github.com/allenai/allennlp-models}{https://github.com/allenai/allennlp-models}) for coreference resolution. We used the spacy toolkit (\href{https://github.com/huggingface/neuralcoref}{https://github.com/huggingface/neuralcoref}) to identify named entities in the text.

\section{Evaluation Metrics}
\label{sec:metrics}
\subsection{Automatic Evaluation:} For evaluating our baseline and proposed models, we used F1\footnote{https://github.com/facebookresearch/ParlAI/blob/master/
parlai/core/metrics.py}, BLEU \citep{papineni2002bleu}, PPL and Embedding-based metrics\footnote{https://github.com/Maluuba/nlg-eval} \citep{liu2016not} such as Vector Extrema, Greedy Matching and Embedding Average for evaluation. Perplexity (PPL) is a metric used to assess how well a probability model predicts a sentence. The term intersection between the gold response and output response by the model is calculated using BLEU (BLEU-4) and the unigram F1-score. Word-matching-based metrics are an alternative to embedding-based metrics. These metrics allocate a vector to each term in the sentence in order to truly understand the intended meaning of the predicted sentence, as described by word embedding. Using the above standard metrics, we evaluate our models on both the seen and unseen test sets of the Wizard of Wikipedia dataset, as well as the test set of the CMU\_DoG dataset.

\subsection{Human Evaluation:} However apart from the automatic evaluation metrics, for evaluating samples from human perspective we randomly selected 100 samples from the Wizard of Wikipedia's Test Seen and Test Unseen sets. We hire four professionals, each with a post-graduate degree and experience, to serve as human judgment annotators. The annotators are regular employees (paid monthly in accordance with university policy) earning Rs 35,000 per month. The annotators are members of our research team and have been working on similar projects for the past three years. For each example, we provide our annotators with model responses and human ground-truth.
We use the following metrics for evaluation:

\begin{inparaenum}[(i)] 
\item Fluency: It is a metric that measures whether or not a sentence is comprehensible.
\item Adequacy: This metric assesses the cohesiveness of the generated response with respect to the conversation context.
\item Knowledge Existence (KE): This metric determines whether the response contains knowledge or not.
\item Knowledge Correctness (KC): This metric determines whether the knowledge in the predicted response is correct.
\item Knowledge Relevance (KR): This metric is used to determine whether the knowledge is correct and relevant to the topic of the conversation.
\end{inparaenum}
The annotators assign a score of 0 to 2 to each response (representing "incorrect," "moderately correct," and "perfect"). Fleiss' kappa \citep{fleiss1971measuring} is used to calculate the annotators' agreement.

\section{Results}

\subsection{Automatic Evaluation}
\label{sec:auto_r}
We also compare our proposed CNTF model to \citep{zhao2020multiple} and \citep{zheng-etal-2021-knowledge}. MKST \citep{zhao2020multiple} obtains a F1-score of 22.2 / 21.3 and BLEU-4 score of 0.077 / 0.072 on test seen / unseen of WoZ dataset. KTWM \citep{zhao2020multiple} obtains a BLEU-4 score of 0.033 / 0.022 with an embedding average, extrema and greedy score of 0.682 / 0.668, 0.394 / 0.379, 0.574 / 0.542 respectively. Our model clearly outperforms these baselines by obtaining a BLEU-4 score of 0.119 / 0.110, F1-score of 32.5 / 31.4 with an embedding average, extrema and greedy score of 0.911 / 0.910,  0.577 / 0.570, 0.758 / 0.752, respectively. In addition, our model clearly outperforms the BART based models for knowledge grounded generation \citep{de2020bart} on F1-score (Test Seen - 12.2 / 20.1; Test Unseen 14.9 /19.3) by a huge margin on both the test set of WoZ dataset.

\subsection{Error Analysis}
\label{sec:error_ext}
{For a dialogue with no topic specific knowledge sentences usually our model fails to keep the conversation going by generating inadequate responses and also misses several entities. For example, \textit{Input utterance}: that's not uncommon! there are rescue groups that specialize in finding homes for retired sled dogs. I bet they retire them at a certain age then they need a home huh; \textit{Predicted Response (CNTF)}: that's cute! i'm sure they're cute!; \textit{Gold Response}: yes. huskies got their name from the word referring to eskimos. As it can be clearly seen the model fails to capture the entity \textit{huskies} and instead generates a generic response.}

\end{document}